\documentclass[letterpaper, 10 pt, conference]{ieeeconf}
\usepackage{graphicx} 
\usepackage{hyperref}
\usepackage{subcaption}
\usepackage{colonequals}
\usepackage{amsmath} 
\usepackage{float}
\usepackage{booktabs}
\usepackage{amsfonts}
\usepackage{amssymb}
\usepackage[dvipsnames]{xcolor}
\usepackage{siunitx}
\usepackage[normalem]{ulem}
\usepackage[font=small,labelfont=bf]{caption}

\usepackage[ruled]{algorithm2e}
\makeatletter
\newcommand{\algorithmstyle}[1]{\renewcommand{\algocf@style}{#1}}
\newcommand{\removelatexerror}{\let\@latex@error\@gobble}
\newcommand{\norm}[1]{\left\lVert#1\right\rVert}
\usepackage[noadjust]{cite}

\makeatother

\IEEEoverridecommandlockouts                              

\def \kkk{\color{black}}

\title{\LARGE \bf
	A Geometric Perspective on Visual Imitation Learning
}

\author{Jun Jin$^{\dagger}$, Laura Petrich$^{\dagger}$, Masood Dehghan$^{\dagger}$ and Martin Jagersand$^{\dagger}$
\thanks{$^{\dagger}$Authors are with Department of Computing Science,
        University of Alberta, Edmonton AB., Canada, T6G 2E8.
        { 
           \tt\small \{jjin5, laurapetrich, masood1, mj7\}@ualberta.ca
        }
        }%
}

\begin{document}
	\maketitle
	\thispagestyle{empty}
	\pagestyle{empty}
	
	\begin{abstract}
		We consider the problem of visual imitation learning without human supervision (e.g. kinesthetic teaching or teleoperation), nor access to an interactive reinforcement learning (RL) training environment. We present a geometric perspective to derive solutions to this problem. Specifically, we propose VGS-IL (Visual Geometric Skill Imitation Learning), an end-to-end geometry-parameterized task concept inference method, to infer globally consistent geometric feature association rules from human demonstration video frames. We show that, instead of learning actions from image pixels, learning a geometry-parameterized task concept provides an explainable and invariant representation across demonstrator to imitator under various environmental settings. Moreover, such a task concept representation provides a direct link with geometric vision based controllers (e.g. visual servoing), allowing for efficient mapping of high-level task concepts to low-level robot actions.
	\end{abstract}
	
	\section{Introduction}
	\label{sec:intro}
	Compared to traditional robotic task teaching methods, visual imitation learning promises a more  intuitive way for general purpose task programming. Like most other learning methods, it suffers from the generalization problem.
	Commonly, three strategies are used to tackle generalization. The first one is to increase the number of  human demonstrations via kinesthetic teaching or teleoperation.  This has been proven effective for supervised learning methods  such as behavior cloning~\cite{argall2009survey}. However, it requires laborious human supervision which can be tedious. The second  strategy is to assume access to robot-environment interactions where samples can be expanded via reinforcement learning (RL) methods (e.g. IRL~\cite{Abbeel2004}, GCL~\cite{finn2016guided}, GAIL~\cite{Ho2016}). Unfortunately, new issues regarding transfer learning and low sample efficiency arise during both simulation and real world training. The last strategy assumes that shared knowledge can be learned from  demonstration samples across multiple but similar tasks; from this shared knowledge, the robot is able to learn a new task when given one more demonstration. This strategy is used in meta-learning based approaches (e.g. one-shot~\cite{finn2017one}).
	
	\begin{figure}
		\setlength{\belowcaptionskip}{-10pt}
		\begin{center}
			\includegraphics[width=0.5\textwidth]{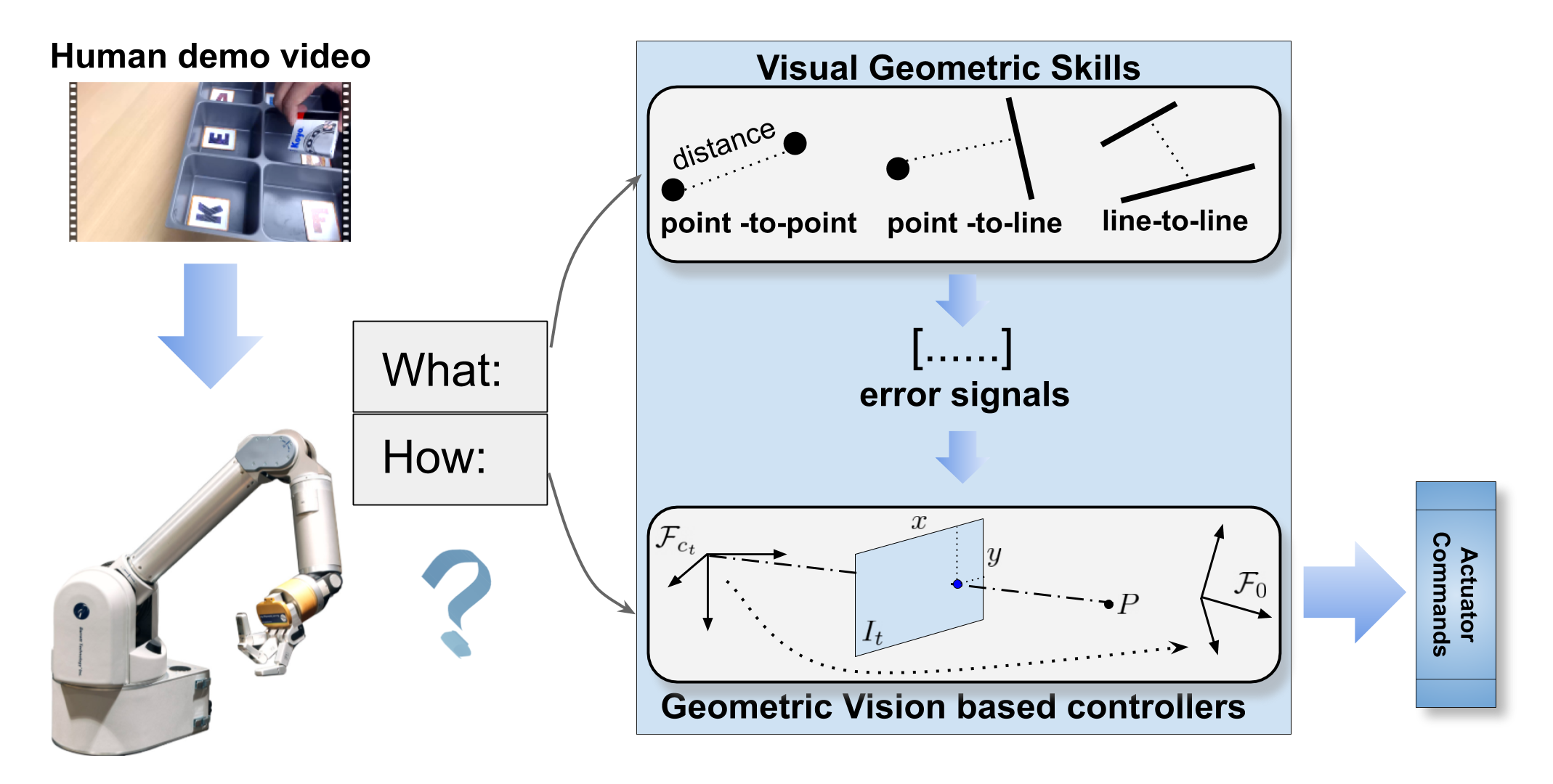} 
			\caption{Rethinking the classical `correspondence' problem~\cite{Argall2009} in imitation learning reveals two essential questions: i) `\textit{what}' information should be transferred from a human demonstrator to a robot imitator; and ii) `\textit{how}' can this information be used to bring about actions (i.e. motor action). 
			This paper presents a geometric perspective to this problem.  We find a proper geometric  representation of `\textit{what}' that facilitates the training of `\textit{how}'. 
			}
			\label{fig:design_overview}
		\end{center}
	\end{figure}
	
	Generally, aforementioned methods  use human demonstration as \textit{state-action} samples to learn a policy mapping from image to action (i.e. they approximate a target state-action distribution). Consequently, in order to improve generalization, it is necessary to collect  more state-action experiences from either human teaching (supervision) or robot self-explorations (RL training). However, neither approach proves satisfactory. This motivates us to ask the question: \textit{is it possible to  learn by watching \textbf{one} human demonstration without the extra effort of interactive training?}
	
	Recently, several methods have been proposed to tackle this question. One \textit{key insight} is to rethink the classical `correspondence' problem~\cite{Argall2009} which studies the difference between demonstrator and imitator. Such insight changes our view on human demonstration  towards  encoding more task concepts rather than control. Empirically, this aligns with our cognitive process in peer learning which involves first understanding the task before attempting any motor actions\footnote{This is studied in observational learning~\cite{burke2010neural} in psychology.}. This hierarchical view decouples learning the `what' and `how' (zero-shot~\cite{pathak2018zero}, see Fig. 1). Benefits of this method are immediately observed: i) the promise to generalize well since it learns a high-level cognitive concept~\cite{jin2018robot,sharma2019third} of the task instead of directly matching state-action distributions; and ii) the promise of reusable low level polices as basic skills across different tasks~\cite{jin2019visual}. However, two new problems arise: i) what is the high-level task; and ii) how  can we train the low level controllers without an additional intensive cost.
	
	In this paper, we provide a geometric perspective to derive solutions. We show that, instead of learning from image pixels to actions, learning a geometry-parameterized task concept\footnote{For further reading, task parameterization using geometric constraints (e.g.  point-to-point, point-to-line, point-to-conics, etc.) are intensively studied in~\cite{dodds1999task,Dodds1999a,Dodds1999,Hager2000}.} provides an explainable and invariant representation across demonstrator to imitator under various environmental settings. Moreover, it provides \textit{controllability} that  can be directly linked to geometric vision based controllers (e.g. visual servoing). Our contributions are:
	\begin{itemize}
		\item We propose VGS-IL (Visual Geometric Skill Imitation Learning), an end-to-end geometry-parameterized task concept inference method  used to infer globally consistent geometric association rules from demonstration video frames. Instead of learning from geometric primitives~\cite{jin2019visual,ahmadzadeh2015learning} (e.g. points, lines, and conics) with handcrafted feature descriptors~\cite{lowe1999object,rublee2011orb,zhang2013efficient}, VGS-IL  can directly optimize a combinatorial representation from image pixels. Experiments show that the learned task concept generalizes well from human to robot despite the visual difference in arm and hand appearance.
		\item We show such geometry-parameterized task concept can be directly linked to geometric vision based controllers~\cite{corke1994high}, thus forming an efficient way to map high-level task concept to low-level robot actions. Unlike prevalent methods requiring hierarchically training of an additional control policy~\cite{pathak2018zero,finn2017one,sharma2019third,florence2018dense,florence2019self},  experimental results show that our learned representation fits directly into a visual servoing~\cite{Chaumette2006} controller, removing the need  for feature trackers. 
	\end{itemize}
	
	By using geometric primitive associations and 3D computer vision geometry based controllers,
	we present a  method for general purpose robotic task programming.\kkk
	
	\section{Related Works}
		\label{sec:review}
	\textbf{Visual Imitation Learning}: The problem defined in visual imitation learning is: given one or several human demonstration videos, how  can a new task be learned? Research on this topic dates back to 1994~\cite{ikeuchi1994toward,kuniyoshi1994learning}. With the rise of deep learning and reinforcement learning, more influential works have  since been published. While some are reviewed in section \ref{sec:intro}, which aim to learn a task from visual inputs, it's worth noting another research stream aiming to learn a \textit{semantic knowledge} representation. This method commonly relies on independent pipelines like object detection, action recognition etc. Despite of their method complexity, experiments show they can learn semantic task plans  that follow  a procedural manner~\cite{xiong2016robot,ahmadzadeh2015learning,yang2015robot}.
	
	\textbf{Hierarchical Visual Imitation Learning}: Instead of simultaneously learning task definition and control, hierarchical approaches decouple the two by focusing  on learning a shared high-level task representation across human demonstrator and robot imitator. The two core problems are: i) how to represent the high-level task concept; and ii) how to train the low-level control policy. The first one is more important since representation of the task concept determines the controller training. For example, many pioneer works parameterize the task concept in \textit{pixel level} by using sub-goal output  from  a neural network~\cite{pathak2018zero,sharma2019third}. The low-level policy is then sub-goal conditioned and trained following a Hierarchical Reinforcement Learning manner.
	
	More recent works represent a task in the \textit{object level} where object correspondence~\cite{florence2018dense,florence2019self} or graph structure~\cite{sieb2019graph} relationships are utilized to parameterize a task. The low-level controller is then trained based on distance errors in the embedded parameterization space. This approach shows success in pushing and placing tasks, however, it lacks the  definition resolution required for more complex tasks like insertion. 
	
	\textbf{Geometry-Based Visual Imitation Learning}: Alternatively, going deeper inside the objects, \textit{geometric feature level} based approaches arise. Early pioneer works from Ahmadzadeh et al. 2015~\cite{ahmadzadeh2015learning} proposed VSL to learn feature point correspondence based task representation given one human demo video. A similar approach from Qin et al. 2019~\cite{qin2019keto} presents KETO which utilizes key point relationships to represent a tool manipulation task. In general, their low-level controllers are tediously trained separately without enough study emphasis on how a proper task representation will facilitate the low-level policy training.
	
	Beyond a simple key point correspondence based task concept representation, other basic geometric constraints (point-to-line, line-to-line, etc.) can enrich our toolbox for parameterization of  task concepts~\cite{Gridseth2016}. Furthermore, by concurrently combining and sequentially linking them~\cite{Dodds1999}, we can find a general way to program more complex manipulation tasks that exhibit scalability. To the authors best knowledge, applying such systematic geometry-based task programming in visual imitation learning is rarely studied.
	
	\begin{figure*}
	\setlength{\belowcaptionskip}{-10pt}
		\centering
		\includegraphics[width=0.98\textwidth]{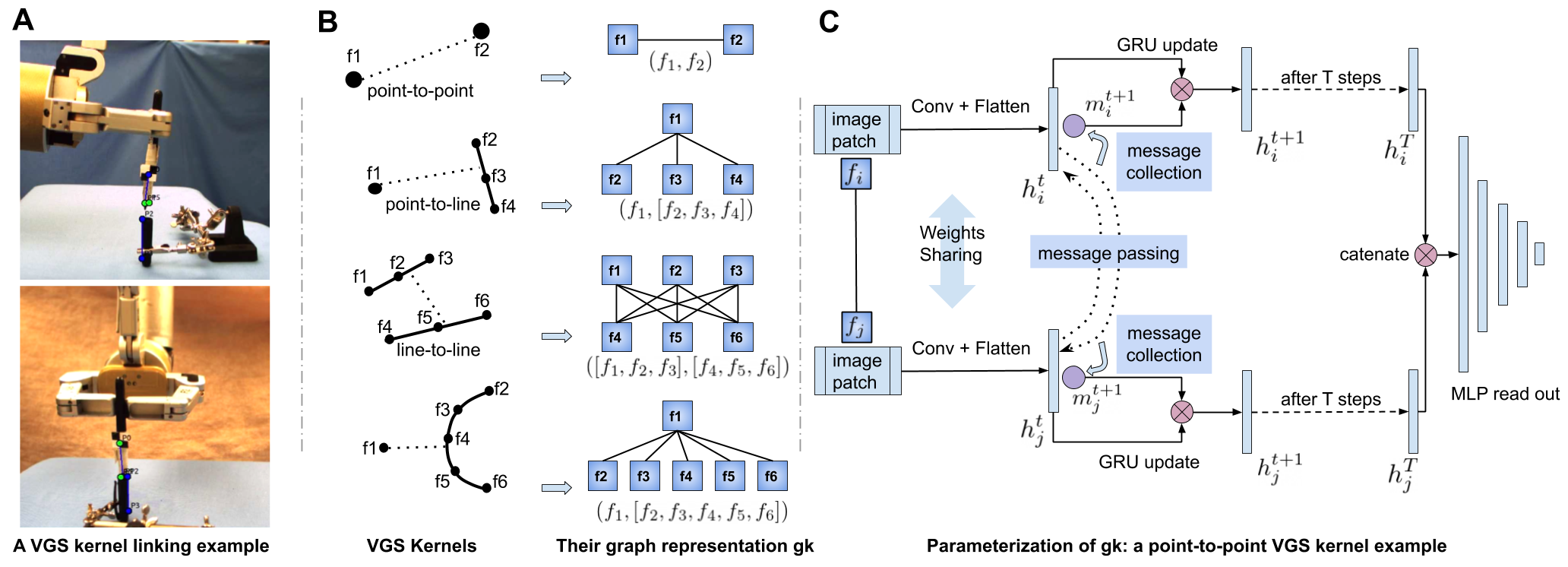}
		\caption{\textbf{A}: An example of visual geometric skill (VGS) kernel linking task~\cite{gridseth2016vita}: Inserting a pen inside its cap involves a point-to-point constraint (pen tip touches cap top) followed by a line-to-line constraint (align pen direction to the cap). \textbf{B}: Basic geometric skill kernel representations using a graph structure. More complex tasks are programmed by combining the basic kernels. \textbf{C}: An example of parameterization design of the point-to-point kernel $\mathit{gk}_{p2p}$. Using a message passing graph neural network with a GRU update fulfills the three required properties for a good $\mathit{gk}$ representation. Other visual geometric skill kernels can be  parameterized in a similar manner.}
		\label{fig:diagram}
	\end{figure*}
	
\section{Method}
	\label{sec:method}
	This research builds upon our previous work on visual geometric skills learning~\cite{jin2019visual}, employing a more data driven approach to learn globally consistent geometric feature association rules without hand crafted feature descriptors~\cite{lowe1999object,rublee2011orb,zhang2013efficient}.

	\subsection{Geometry parameterized task representation} \label{sec:geo_task}
	The basic idea of geometric feature association rules based task parameterization, as firstly proposed in~\cite{dodds1999task}, has two parts: i) the basic geometry constraints or visual geometric skill (VGS) kernels~\cite{jin2019visual}; and ii) the combination or conditioned linking of basic kernels to create more complex tasks, which we refer to as visual geometric skills. For example, some commonly used geometric skill kernels are:
	\begin{itemize}
		\item \textit{point-to-point} $\mathit{gk_{p2p}}$: the coincidence of two points.
		\item \textit{point-to-line} $\mathit{gk_{p2l}}$: a point fits on (touches) a line.
		\item \textit{line-to-line} $\mathit{gk_{l2l}}$: a line is co-linear with another line.
		\item \textit{point-to-conic} $\mathit{gk_{p2c}}$: a point fits a conic.
	\end{itemize}
	
	This parameterization method provides a programmable framework that can be used to create more complex tasks by combining several constraints in parallel and then sequentially linking the basic kernels. For example, inserting a pen tip inside its cap involves a point-to-point constraint linked by a line-to-line one (Fig. 2A).

	First, \textbf{how do we define a \textit{good} VGS kernel representation?} Given a set of geometric features $\{f_{i}\}$, a VGS kernel representation is an operator $\mathit{gk}$ used to map $\{f_{i}\}$ to a latent vector. 
	We propose three essential properties of a good $\mathit{gk}$:
	\begin{itemize}
		\item \textit{Communicative:} A good operator $\mathit{gk}$ should be consistent for all input features sequence orders. For example, when we enumerate all possible associations of three features, we require $\mathit{gk}(f_{1}, f_{2}, f_{3})=...=\mathit{gk}(f_{3}, f_{2}, f_{1})$ for all 6 possible permutations since they define the same task.
		\item \textit{Non-inner-associative}: A good $\mathit{gk}$ should be able to represent $\mathit{gk}(f_{1}, [f_{2}, f_{3}, f_{4}])\neq \mathit{gk}([f_{1}, f_{2}, f_{3}], f_{4})$. For example, consider a point $f_{1}$ and a line defined by three points $(f_{2},f_{3},f_{4})$. A point-to-line kernel operation is $\mathit{gk}(f_{1}, [f_{2}, f_{3}, f_{4}])$, which is unique from any other inner associations. 
		\item \textit{Scalability:} The ways to parameterize $\mathit{gk}$ should be scalable to fit n-ary operations. For example, a point-to-point kernel is a binary operation while point-to-line is a quaternary operation if three points represent a line.
	\end{itemize}
	Examples of basic VGS kernel parameterizations are included in Fig. 2B.
	As shown in Jun et. al ~\cite{jin2019visual}, a parameterization is found using a message passing graph neural network~\cite{gilmer2017neural} with a gated recurrent unit GRU~\cite{chung2014empirical} that satisfies the above properties: i) a graph structure is scalable to represent n-ary operations; ii) graph edges define different inner associations; and iii) the message passing mechanism combined with GRU makes $\mathit{gk}$ invariant from input orders. Specifically, this design (Fig. 2C) has four steps: A pair-wise message generation $\mathcal{M}$:
	\begin{equation}
	\begin{aligned}
	m^{t+1}_{i \rightarrow j}=\mathcal{M}(h^{t}_{i}, h^{t}_{j})
	\end{aligned}
	\end{equation} 
	, where $h^{t}_{j}$, $h^{t}_{j}$ are connected nodes' hidden states.
	A message aggregation $\mathcal{A}$ which collects all incoming messages:
	\begin{equation}
	\begin{aligned}
	m^{t+1}_{i}=\mathcal{A}(m^{t+1}_{i \rightarrow j})
	\end{aligned}
	\end{equation} 
	A message update $\mathcal{U}$ using a gated recurrent unit (GRU):
	\begin{equation}
	\begin{aligned}
	h^{t+1}_{i}=\mathcal{U}(h^{t}_{i},m^{t+1}_{i})
	\end{aligned}
	\end{equation}
	Finally a readout function is parameterized using MLP layers. After T layer updates, all nodes' final states are fed into a readout function: $b=\text{MLP}(h^{T}_{1},...,h^{T}_{n})$.
	
	Next, we show \textbf{how to encode the graph entities.} 
	Previous works~\cite{jin2019visual,Dillmann2004} used hand crafted feature descriptors in training. Is it possible to utilize the representation capability of deep learning by directly learning from raw images? Moreover, for simple geometric features like points and lines, there are on-the-shelf descriptors that can be used. What about more complex geometric primitive like conics and planes? In this paper, we propose a composable graph structure used to encode geometric primitives with point-based image patches, as shown in Fig. 2B.
	
	Lastly, a visual geometric skill (VGS) is composed by combining or linking multiple VGS kernels. This paper will only cover kernel combinations and will leave kernel linking for future research.
	
	\subsection{VGS learning by watching human demonstrations}
	Assume a VGS task consists of multiple geometric skill kernels $\{\mathit{gk}_{i}\}$, learning VGS becomes the optimization of each ${\mathit{gk}_{i}}$ given a human demonstration image sequence $\{I_{t}\}$. 
	\subsubsection{Select-out function.} An optimal ${\mathit{gk}_{i}}$ selects the \textit{right} geometric feature associations out of a set of combinatorial instances. For example, in the point-to-point kernel, we can get \textit{N} feature points from one image by applying any feature extractor. To enumerate, there are $m=C^{2}_{N}$ candidate instances. Suppose each instance has an output $b_{j}$ by applying the operator $\mathit{gk_{p2p}}$. We compute its relevant factor $\mathit{g}_{j}=\text{softmax}(b_{j},\{b_{1},...,b_{m}\})$ and the right one is selected out from the maximum $\mathit{g}_{j}$.
	
	\subsubsection{Optimization} Applying ${\mathit{gk}_{i}}$ on each image frame $I_{t}$ will output a control error signal $\mathbf{e}_{t}$\footnote{For example, a point-to-point kernel outputs x-y errors in image pixels. A point-to-line kernel outputs error signal from the dot product of their homogeneous coordinates. More examples can be found in~\cite{hartley2003multiple}.}. Assuming ${\mathit{gk}_{i}}$ is optimal, applying ${\mathit{gk}_{i}}$ on a human demonstration image sequence $\{I_{t}\}$ will output a high-quality control error signal sequence $\{\mathbf{e}_{t}\}$. Hence, optimizing ${\mathit{gk}_{i}}$ is essentially selecting the right observation space to observe a high-quality control error signal output from the human demonstrator. We call this the \textbf{\textit{observational expert}} assumption. By maximizing the quality of control error signals, we are able to adjust our ${\mathit{gk}_{i}}$ estimation (Fig. 3).
	
	We measure the quality of control signals by a reward function using two metrics: i) errors are overall decreasing along the time steps of human demonstration; and ii) error changes are smooth. The first metric is encoded into the reward function as defined in~\cite{jin2019visual}. To achieve smoothness, we modify the loss function defined in~\cite{jin2019visual} by adding a geometry consistent regularizer (GCR): $-\alpha \norm{b_{t+1} - b_{t}}_{2}^{2}$ while keeping the same residual sum of weights (RSW) regularizer for deterministic selection purpose. GCR forces learning a more consistent selection across frames.
	
	\begin{figure}
		\setlength{\belowcaptionskip}{-10pt}
		\begin{center}
			\includegraphics[width=0.45\textwidth]{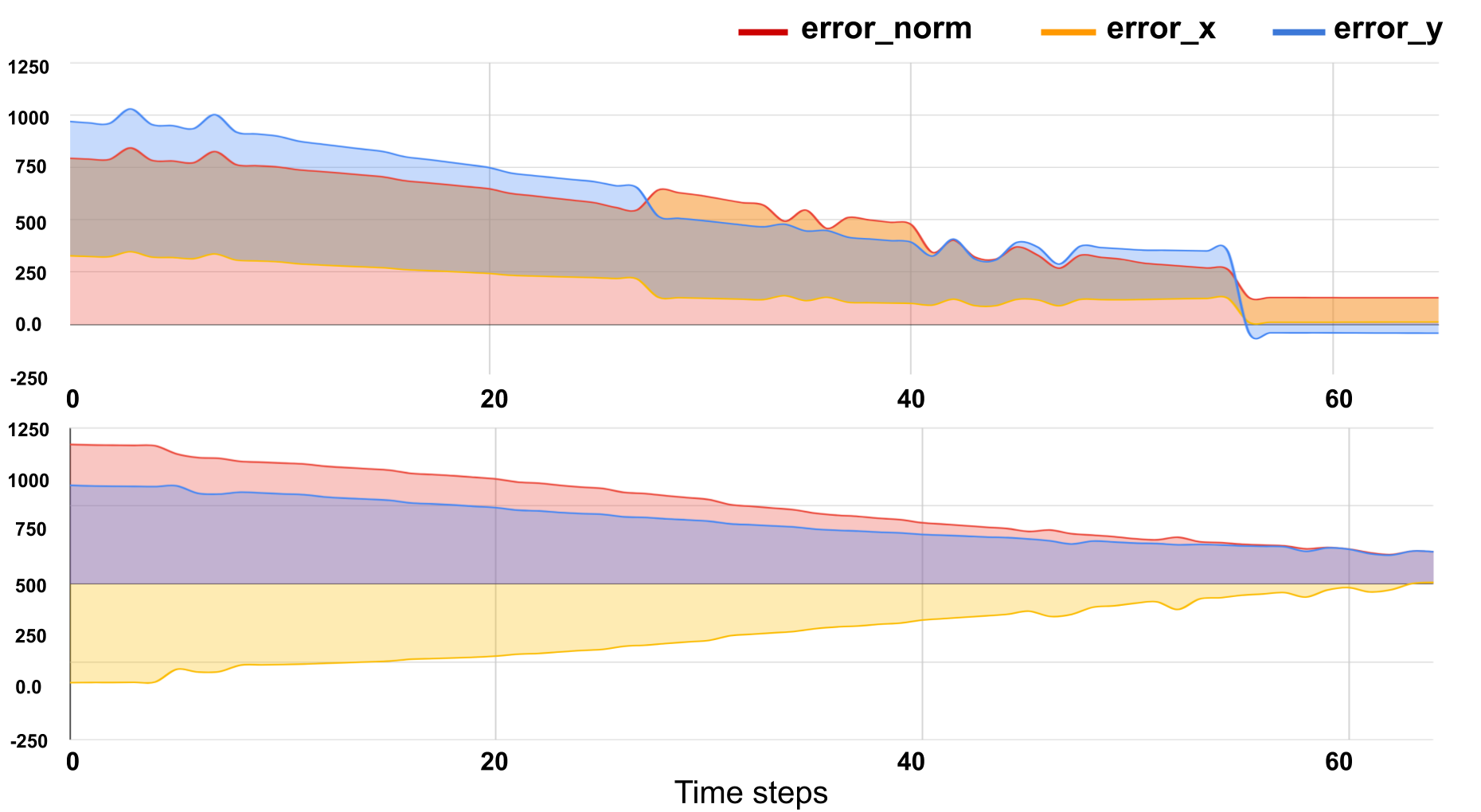} 
			\caption{Given human demonstration video frames, optimizing $\mathit{gk}_{i}$ is essentially selecting the right observation space to measure a high-quality control error signal output from the human expert demonstrator. By maximizing the quality of control error signals, we adjust our $\{\mathit{gk}_{i}\}$ estimation. \textbf{Up}: The control error output of $\mathit{gk}_{i}$ trained without geometry consistent regularizer (GSR) of the \textit{sorting} task described in Sec. \ref{sec:experiments}. \textbf{Down}: The control error output of a well-trained $\mathit{gk}_{i}$ by adding GSR to the loss function.}
			\label{fig:setup}
		\end{center}
	\end{figure}
	
	By optimizing the reward function using InMaxEntIRL~\cite{jin2018robot,jin2019visual}, the control signal quality from the human demonstrator is optimized, resulting in an optimized ${\mathit{gk}_{i}}$. To summarize, we propose VGS-IL (Visual Geometric Skill Imitation Learning) as detailed in Algorithm 1. 
	
	\begingroup
	\removelatexerror
	{\small
		\begin{algorithm}[tbp]
			\SetAlgoLined
			\KwIn{Expert demonstration video frames \{$I_{1},...,I_{n}$\}, demonstrator confidence level $\alpha$, VGS=\{$gk_{1},...,gk_{m}$\}}
			\KwResult{Optimal weights $\boldsymbol{\theta}_{i} ^{*}$ of $\mathit{gk_{i}}$}
			{\textit{Construct kernel graph instances on each frame}}\\
			\For{i=1:m}{
				Define $\mathcal{S}=\{\}$\\
				\For{t = 1:n}{
					Feature extraction on $I_{t}$ according to $gk_{i}$ defined in Section \ref{sec:geo_task}\\
					$s_{t}\leftarrow$ Construct all $gk_{i}$ instances by feature association\\
					$\mathcal{S}\leftarrow$ Append $s_{t}$\\
				}
				{\textit{Prepare State Change Samples $\mathcal{D}{s}=\{s_{t}\rightarrow s_{t+1}\}$}}\\
				$\boldsymbol{\theta}_{i} ^{*}$ = InMaxEntIRL($\mathcal{D}{s}, \alpha, \mathit{gk_{i}}$) \cite{jin2019visual}\\
		    }
			\caption{VGS-IL}
		\end{algorithm}
	}
	\endgroup
	
		
	\subsection{Links to geometric vision-based controllers}
	The control signal $\mathbf{e}_{t}$ output from ${\mathit{gk}}$ is observed in image pixel space. Mapping image observations to robot actions is a long running research topic~\cite{agin1977servoing} also known as robot eye-hand coordination, visuomotor policy learning, or vision guided robot control~\cite{corke1994high}. Approaches can be divided into two  categories\footnote{For further reading, a comparison has been discussed in the ICRA 2018 Tutorial on Vision-based Robot Control~\cite{chaumette2018geometric}}: i) end-to-end learning methods~\cite{Levine2016}; and ii) visual servoing (VS)~\cite{Chaumette2006}. 
	End-to-end learning approaches can work without explicit features, and are useful in complex visual environments due to their powerful representation capability~\cite{Lecun2015}, but require time consuming training and show poor transfer to new environments (i.e. poor generalization).
	Visual servoing approaches run in real-time using a geometric vision-based control law, but can lack sufficient visual representation capability. 
	Combining the geometric vision part from visual servoing with learning-based methods is rarely studied~\cite{bateux2018going,bateux2018training}.

	\begin{figure}[tbp]
		\setlength{\belowcaptionskip}{-10pt}
		\begin{center}
			\includegraphics[width=0.4\textwidth]{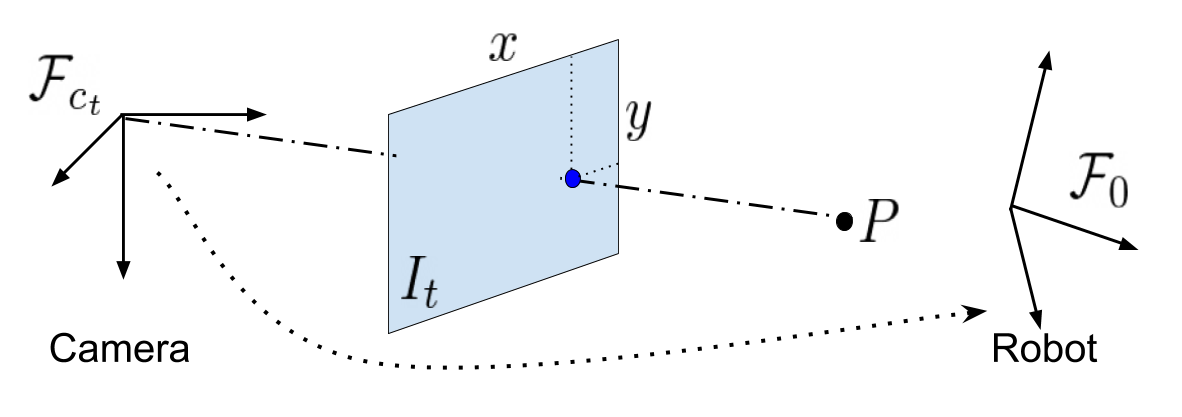} 
			\caption{A geometric vision-based controller utilizes camera 3D geometry to build relationships between 3D object motion, observed feature motion in image plane $I_{t}$ and camera spatial velocity. At last it links with robot actions via a calibration model or trial-error manner based online learning.}
			\label{fig:baseline}
		\end{center}
	\end{figure}

	As shown in Fig. 4, the basic idea of using geometric vision in VS control is: i) mapping an error vector $\dot{\mathbf{e}_{t}}$ to camera motion $\mathbf{v}_{c_{t}}$ via an interaction matrix derived from the camera relative spatial velocity equation~\cite{Chaumette2006}; and ii) mapping $\mathbf{v}_{c_{t}}$ to robot motion $\mathbf{a_{t}}$ via a calibration model as in VS or a trial-error based online estimation as shown in Uncalibrated Visual Servoing (UVS~\cite{Jagersand1997}). Here we discuss feature-based visual servoing which our VGS learning directly links to. 
	
	VGS-IL removes the need for robust feature trackers while keeping the geometric error output  that can be linked with  a visual servoing controller. Compared to traditional approaches that hand select features to encode a task concept, VGS-IL directly learns the  feature selection using a data driven approach. Instead of tracking each geometric feature and then associating them, VGS-IL directly extracts their associations in an adaptive manner which has been shown to be more robust~\cite{jin2019visual}.
	
	It is worth noting that visual servoing control is sensitive to modeling errors~\cite{chaumette2018geometric}. Combining the 3D geometric vision aspect from VS to learn more robust controllers via Reinforcement Learning has the potential to derive both efficient and robust controllers.
	
	\begin{figure*}
		\centering
		\includegraphics[width=0.98\textwidth]{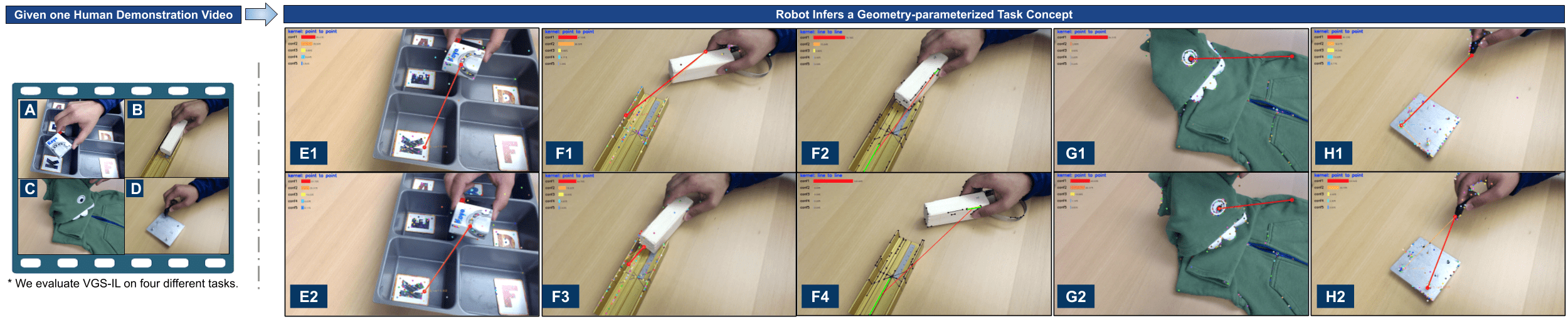}
		\caption{\textbf{Left}: Four tasks designed in evaluation: \textit{Sorting}, \textit{Insertion}, \textit{Folding} and \textit{Screw}. \textbf{Right}: Qualitative evaluation results of VGS-IL in the four tasks. We select two frames for each task. The \textit{Insertion} task includes two columns representing point-to-point and line-to-line kernel respectively. For a fair test, we changed the background and target pose in each task. Red line indicates selected feature association with highest confidence. Experiments show VGS-IL succeeds to learn a consistent geometry-parameterized task concept from human demonstrator in all the four tasks. Quantitative results are displayed in Table 1-2 below .}
		\label{fig:diagram}
	\end{figure*}
	
	\begin{table*}[t]
		\begin{center}
		\begin{tabular}{@{}llllcllllll@{}}
			\toprule
			\multicolumn{1}{c}{Task} &
			\multicolumn{2}{c}{\textit{Sorting}} &
			\multicolumn{2}{c}{\textit{Insertion: point-to-point}} &
			\multicolumn{2}{c}{\textit{Insertion: line-to-line}} &
			\multicolumn{2}{c}{\textit{Folding}} &
			\multicolumn{2}{c}{\textit{Screw}} \\ \midrule
			\multicolumn{1}{c}{Metrics} &
			\multicolumn{1}{c}{Acc} &
			\multicolumn{1}{c}{conAcc} &
			\multicolumn{1}{c}{Acc} &
			conAcc &
			Acc &
			conAcc &
			Acc &
			conAcc &
			Acc &
			conAcc \\ \midrule
			Baseline1 & 100.0\% & 1.00 & 100.0\%  & 1.00  & 100.0\% & 1.00 & 10.0\% & n/a & 8.2\% & n/a \\
			Baseline2 & 100.0\% & 0.03 & 100.0\% & 0.02 & 81.2\% & -0.06 & 80.0\% & 0.10 & 33.0\% & 0.08 \\
			VGS-IL      & 100.0\% & 0.98 & 100.0\% & 0.85 & 93.0\% & 0.91 & 84.0\% & 0.98 & 49.0\% & 0.92 \\ \bottomrule
		\end{tabular}
		\caption{Quantitative evaluation of VGS-IL in the four tasks. All tests are based on changed background and randomly placed target. Results show VGS-IL performs better in learning a consistent geometry-parameterized task concept. }
		\label{tab:expA}
	\end{center}
	\end{table*}
	\begin{table*}[t]
		\begin{center}
			\begin{tabular}{@{}llllcllllll@{}}
				\toprule
				\multicolumn{1}{c}{Settings} &
				\multicolumn{2}{c}{\textit{Random Target}} &
				\multicolumn{2}{c}{\textit{Change Camera}} &
				\multicolumn{2}{c}{\textit{Object Occlusion}} &
				\multicolumn{2}{c}{\textit{Object Outside FOV}} &
				\multicolumn{2}{c}{\textit{Change Illumination}} \\ \midrule
				\multicolumn{1}{c}{Metrics} &
				\multicolumn{1}{c}{Acc} &
				\multicolumn{1}{c}{conAcc} &
				\multicolumn{1}{c}{Acc} &
				conAcc &
				Acc &
				conAcc &
				Acc &
				conAcc &
				Acc &
				conAcc \\ \midrule
				Baseline1 & 100\% & 1.00 & 0.0\%  & n/a  & 0.0\% & n/a & 0.0\% & n/a & 0.0\% & n/a \\
				Baseline2 & 99.1\% & -0.03 & 96.7\% & -0.10 & 92.7\% & -0.05 & 81.2\% & -0.03 & 0.0\% & n/a \\
				VGS-IL      & 100.0\% & 0.55 & 95.0\% & 0.61 & 97.3\% & 0.10 & 79.8\% & 0.19 & 19.2\% & 0.42 \\ \bottomrule
			\end{tabular}
			\caption{Evaluation results of VGS-IL on the robot imitator under different environmental settings (shown in Fig. 6). We keep testing on the real robot in the \textit{Sorting} task, while exploring more variance settings. Results show VGS-IL performs the best under all conditions.}
			\label{tab:expB}
		\end{center}
	\end{table*}

    \section{Experiments}
	\label{sec:experiments}
	Through experimental evaluation we aim to determine: (i) whether VGS-IL can learn a correct and consistent geometry-parameterized task concept given one human demonstration; and (ii) whether VGS-IL can output high-quality error signals for  accurate robot  control. For analysis, we decompose the two goals into four evaluation steps: (1) Given one human demonstration video, will VGS-IL output a correct and consistent task concept; (2) how  will VGS-IL generalize from human demonstrator to robot imitator under changed task and environmental settings; (3) How does control error converge, and how is it affected at different network training time for VGS-IL.
	
	\textbf{Baselines:} We hand designed two baselines to use in comparison. \textit{Baseline1} is conventional visual servoing with a video-tracking of a redundant feature set. This involves human interaction to carefully hand select 10 pairs of geometric features used to represent a task and initialize multiple feature trackers for each camera. As long as one pair out of ten is able to track throughout the entire task process, \textit{baseline1} succeeds. \textit{Baseline2} is a method from our previous work \cite{jin2019visual} that relies on hand crafted geometric feature descriptors (SIFT~\cite{lowe1999object} and LBD~\cite{zhang2013efficient}) in training; however, it doesn't take into consideration representation consistency.
	
	\textbf{Metrics:} We designed two evaluation metrics: (1) \textit{Acc} to measure accuracy; and (2) \textit{conAcc} to measure consistency. Specifically, given \textit{N} video frames, \textit{Acc}$=\frac{M\times 100}{N}\%$, where \textit{M} is the number of frames with correct geometric task concept inference. Defining \textit{conAcc} is more challenging since directly measuring the inference consistency involves complex statistical methods~\cite{tarpey1996self}. For simplicity, we measure the time-series control error output $\{\mathbf{e}_{t}\}$ (i.e. the inference outcome) and define \textit{conAcc} $=Autocorr(\{\norm{\mathbf{e}_{t}}\}, k)$, which is the autocorrelation measurement over time-series error norms with \textit{shift=k}. We fix \textit{k=2} in all experiments. Since \textit{baseline1} is a collection of redundant pairs of trackers, measuring the \textit{conAcc} is difficult. In this case we assume that \textit{conAcc=1}, the maximum, if \textit{baseline1} succeeds.
	
	\textbf{Tasks:} To facilitate comparisons, we follow the same four tasks: Sorting, Insertion, Folding, and Screw tasks as defined in~\cite{jin2019visual} (see Fig. 5 for details). \textit{Sorting} represents a rich texture clue task that requires a point-to-point kernel; the \textit{Insertion} task needs a combination of point-to-point and line-to-line kernels; the \textit{Folding} task represents deformable object manipulation; and the \textit{Screw} task has low image textures.

	\subsubsection{Evaluation on human demonstration videos}
	Our first step is to evaluate if VGS-IL learns a both correct and consistent geometric feature associations, given one human demonstration video. For a fair test, we changed both background and target pose in evaluation. Qualitative results are displayed in Fig. 5. Quantitative metric scores are shown in Table \ref{tab:expA}. Results show VGS-IL succeeds to generalize the learned geometry-parameterized task concept in all the four tasks. Regarding selection consistency, VGS-IL performs the best compared to other two baselines.
	
	\subsubsection{Generalization under different environmental settings}
	Then we test if the learned task concept generalizes from human demonstrator to robot imitator. A WAM robot equipped with a Barret Hand is used to test the \textit{Sorting} task (Fig. 7). Furthermore, we keep testing on the robot while exploring more variance settings (Fig. 6): (a) \textit{random target}; (b) \textit{move camera}: 
	We test for real-world projective invariance by randomly translating and rotating the camera; (c) \textit{object occlusion}; (d) \textit{object outside FOV}: The object moves outside the camera's field of view and each method is required to automatically recover when the object is back in the image; and (e) \textit{change illumination}: the lighting condition is changed by adding a spotlight light source. We pick the task \textit{Sorting} to evaluate. Results are shown in Table~\ref{tab:expB} which indicate VGS-IL performs the best in all settings.
	
	\begin{figure}
		\setlength{\belowcaptionskip}{-10pt}
		\begin{center}
			\includegraphics[width=0.5\textwidth]{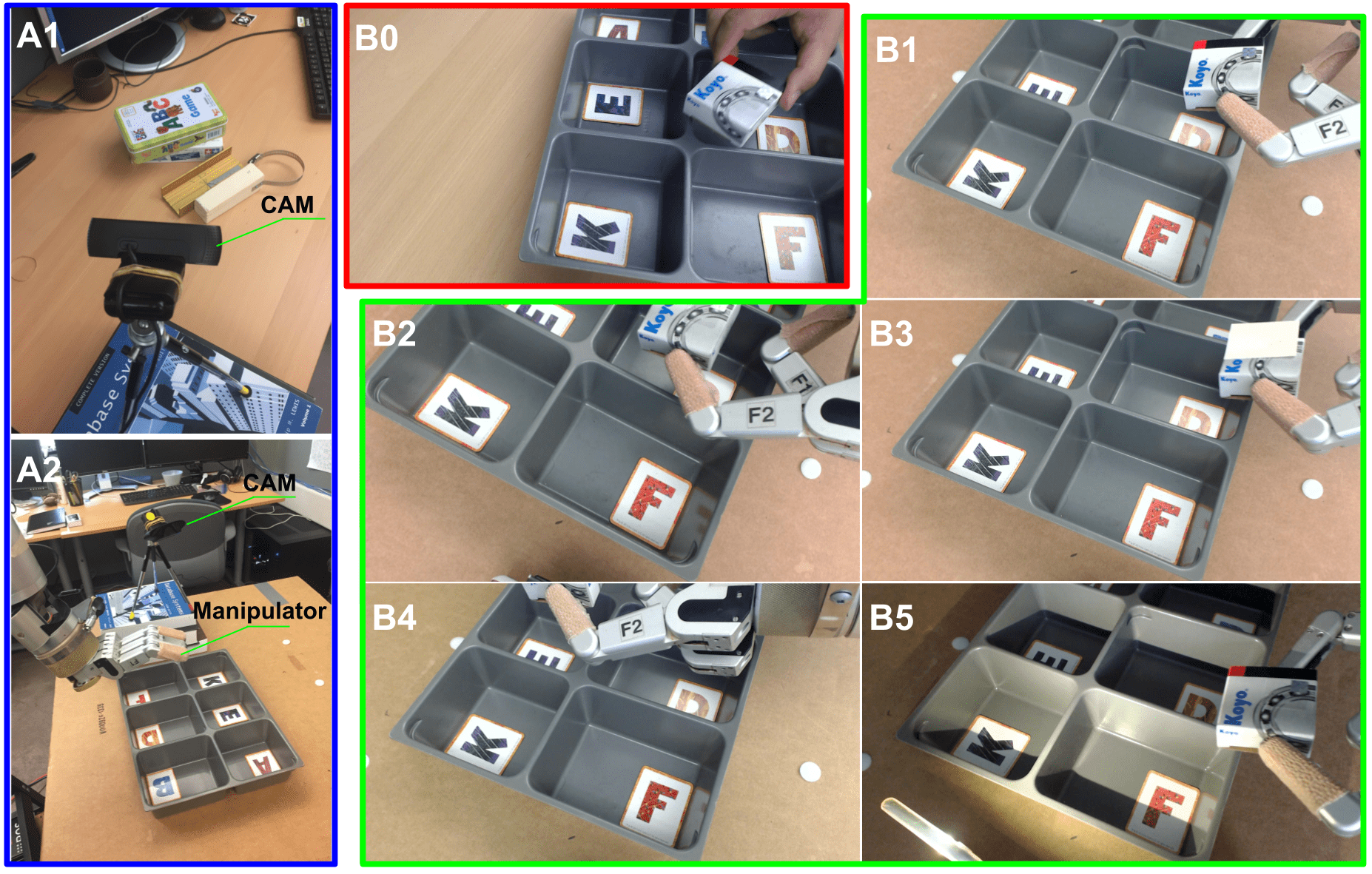} 
			\caption{\textbf{A1}: Human demonstration settings. \textbf{A2}: Robot imitation settings. \textbf{B0}: Human demonstration video used to train VGS-IRL. \textbf{B1-5}: Evaluation on robot under five different environmental settings. B1) \textit{random target}; B2) \textit{change camera}; B3) \textit{object occlusion}; B4) \textit{object outside camera's FOV}; B5) \textit{change illumination}.}
			\label{fig:asf sig}
		\end{center}
	\end{figure}
	
	\begin{figure}
		\setlength{\belowcaptionskip}{-10pt}
		\begin{center}
			\includegraphics[width=0.5\textwidth]{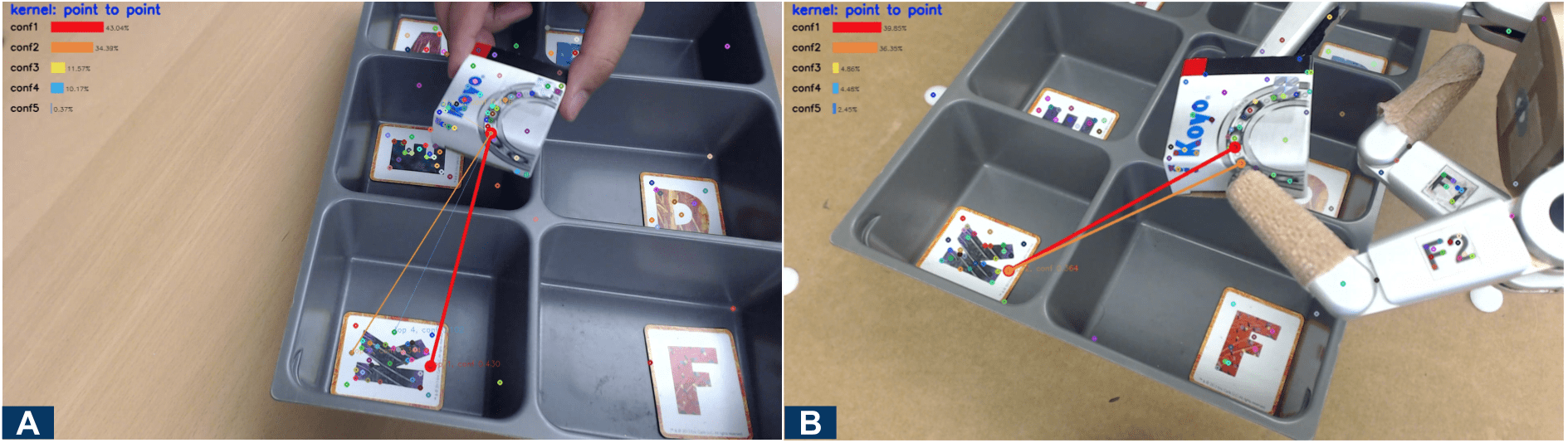} 
			\caption{Example of VGS-IL results in \textit{Sorting} task. \textbf{A}: results in a human demo. \textbf{B}: results in a robot demo. Top five geometric feature associations are selected. Only the top one, as marked red color, is used in evaluation. Results show the same feature point association is selected regardless of human hand or robot hand under different backgrounds and target poses.}
			\label{fig:interface}
		\end{center}
	\end{figure}
	
	\begin{figure}
		\setlength{\belowcaptionskip}{-10pt}
		\begin{center}
			\includegraphics[width=0.5\textwidth]{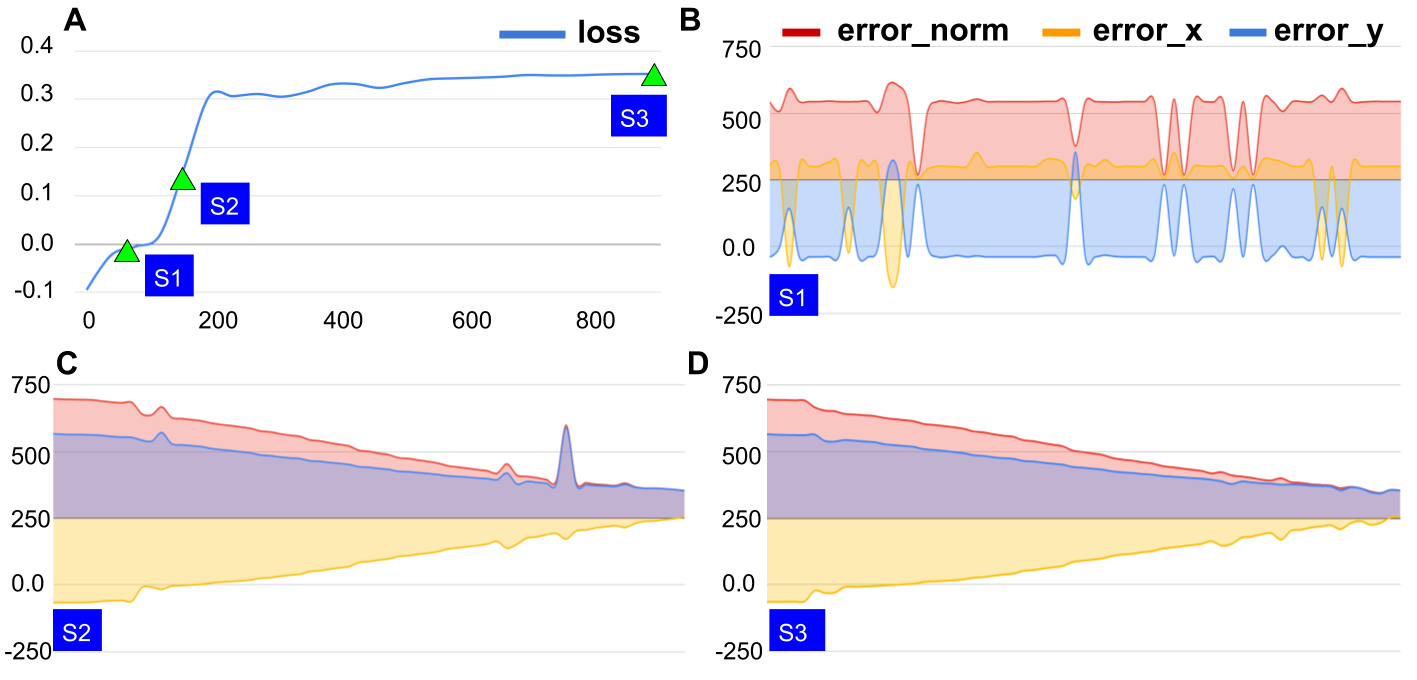} 
			\caption{Evaluation of the control error signals output from VGS-IL in the \textit{Sorting} task. \textbf{A}: Training curve of VGS-IL with three different stages picked for evaluation. \textbf{B, C, D}: Control error signals output from VGS-IL trained in stage \textit{S1, S2, S3}. Results clearly show that VGS-IL outputs a `good' control error signal. Moreover, the optimization process is indeed optimizing the quality of control error signals.}
			\label{fig:control sig}
		\end{center}
	\end{figure}

	\subsubsection{Evaluation of the `good' control error signal output}
	We test how `good' or `bad' the control error signals output from VGS-IL are. To do this, we had the robot perform the \textit{Sorting} task via teleoperation, then ran VGS-IL on the resulting task video and measured the corresponding time-series error signals. Therefore, if VGS-IL was capable of outputting `good' control signals, the results of this video should also be good. To make our evaluation more interesting, we wanted to see how control error signals are improved along with the optimization process of VGS-IL. Fig. 8 shows the results in three different training stages.
	
	\section{Conclusion}
	\label{sec:conclusion}
	We present a geometric perspective on visual imitation learning. Specifically, we propose VGS-IL, visual geometric skill imitation learning, to learn a geometry-parameterized task concept. VGS-IL infers globally consistent geometric feature association rules from human demonstration video. The learned task concept outputs control error signals that can be directly linked to geometric vision based controllers, thus providing an efficient way to map learned high-level task concepts to low level robot actions. Experimental evaluations show that our method generalizes well from human demonstrator to robot imitator under various environmental settings.

	In practice, VGS-IL needs large GPU computation resource due to its optimization over the whole combinatorial feature association candidates. A potential solution is to utilize high dimensional Bayesian Optimization methods~\cite{shahriari2015taking} to directly estimate geometry representation and association parameters from the observation space. Moreover, although we demonstrated applying VGS-IL in tasks by combining different VGS kernels, it is worth further exploring how to sequentially link VGS kernels to program more complex tasks. 
	\addtolength{\textheight}{-2cm}   
	




	\bibliographystyle{IEEEtran}
	\bibliography{IEEEabrv,IEEEexample}
	
\end{document}